\documentclass{article} 
\newcommand{\set}[1]{\lbrace #1 \rbrace}

\newcommand{\pd}[2]{\frac{\partial #1}{\partial #2}}

\newcommand{\p}[2]{p_{#1}^{(#2)}}

\usepackage{nips10submit_e,times}
\usepackage{amsmath,amssymb}
\usepackage{graphicx}

\title{Minimum Probability Flow Learning}

\author{
Jascha Sohl-Dickstein$^{ad1*}$, Peter Battaglino$^{bd2*}$ and Michael R. DeWeese$^{bcd3}$ \\
\small{$^a$Biophysics Graduate Group, $^{b}$Department of Physics, $^{c}$Helen Wills Neuroscience Institute} \\
\small{$^{d}$Redwood Center for Theoretical Neuroscience} \\
\small{University of California, Berkeley, 94720} \\
\small{$^1$\texttt{jascha@berkeley.edu}, $^2$\texttt{pbb@berkeley.edu}, } \\
\small{$^3$\texttt{deweese@berkeley.edu}, \em$^*$These authors contributed equally.}
}

\nipsfinalcopy 

\begin{document}

\maketitle

\begin{abstract}
Fitting probabilistic models to data is often difficult, due to the general intractability of the partition function and its derivatives.
Here we propose a new parameter estimation technique that does not require computing an intractable normalization factor or sampling from the equilibrium distribution of the model.  This is achieved by establishing dynamics that would transform the observed data distribution into the model distribution, and then setting as the objective the minimization of the KL divergence between the data distribution and the distribution produced by running the dynamics for an infinitesimal time. 
Score matching, minimum velocity learning, and certain forms of contrastive divergence are shown to be special cases of this learning technique.
We demonstrate parameter estimation in Ising models, deep belief networks and a product of Student-t test model of natural scenes.  In the Ising model case, current state of the art techniques are outperformed by approximately two orders of magnitude in learning time, with comparable error in recovered parameters. This technique promises to broaden the class of probabilistic models that are practical for use with large, complex data sets.
\end{abstract}

\section{Introduction}
Estimating parameters for probabilistic models is a fundamental problem in many scientific and engineering
disciplines. Unfortunately, most probabilistic learning techniques require calculating the
normalization factor, or partition function, of the probabilistic model in question, or at least calculating
its gradient. For the overwhelming majority of models there are no known analytic solutions,
confining us to the restrictive subset of probabilistic models that can be solved analytically,
or those that can be made tractable using approximate learning techniques. Thus, development
of powerful new techniques for parameter estimation promises to greatly expand the variety of models that 
can be fit to complex data sets.

Many approaches exist for approximate learning, including mean field theory and
its expansions, variational Bayes techniques and a plethora of sampling or numerical
integration based methods \cite{Tanaka:1998p1984,Kappen:1997p6,Jaakkola:1997p4985,haykin2008nnc}.  Of particular interest are contrastive divergence (CD),
developed by Hinton, Welling and Carreira-Perpi\~{n}\'an \cite{Welling:2002p3,Hinton02}, Hyv\"arinen's score matching (SM) \cite{Hyvarinen05} and the minimum velocity learning
framework proposed by Movellan \cite{Movellan:2008p7643,movellan2008cdg,Movellan93}.

\begin{figure}
\begin{center}
\includegraphics[width=1.0\linewidth]{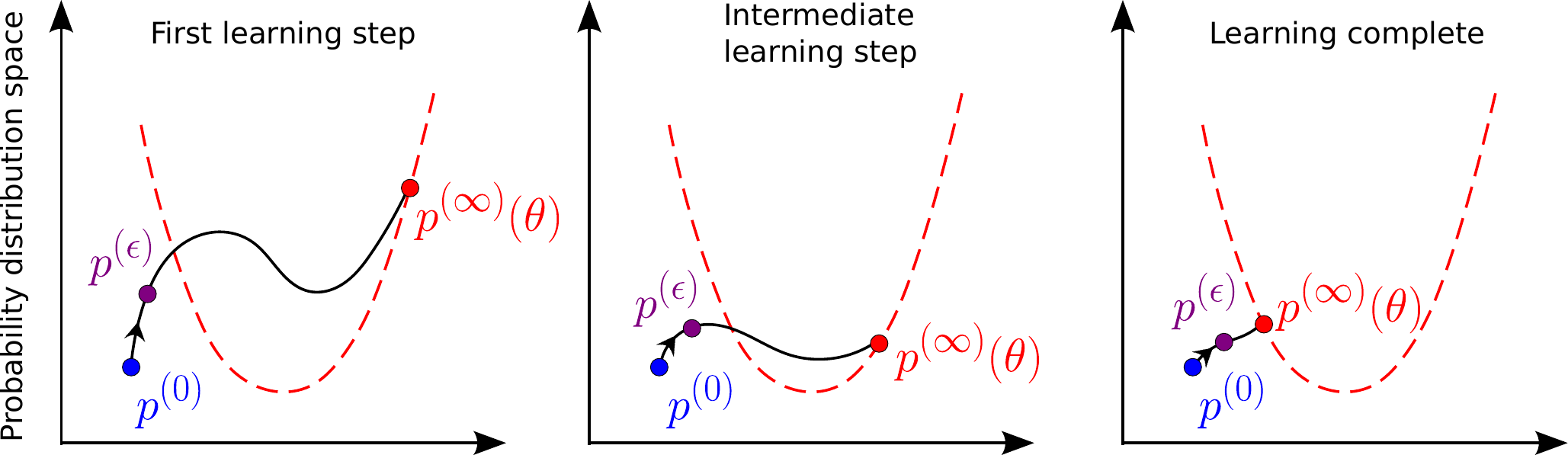}
\end{center}
\caption{
Dynamics are established which transform any initial distribution into the model
distribution $\mathbf{p}^{(\infty)}(\theta)$. The dashed red line indicates the family of
distributions parametrized by $\theta$, and the three successive figures
illustrate graphically the progression of learning. Under maximum likelihood learning, model
parameters $\theta$ are chosen so as to minimize the Kullback--Leibler divergence
between the data distribution $\mathbf{p}^{(0)}$ and the model distribution
$\mathbf{p}^{(\infty)}(\theta)$. Under minimum probability flow learning, however, the KL divergence
between $\mathbf{p}^{(0)}$ and $\mathbf{p}^{(\epsilon)}$ is minimized instead, where
$\mathbf{p}^{(\epsilon)}$ is the distribution obtained by evolving $\mathbf{p}^{(0)}$ for
infinitesimal time under the dynamics. Here we represent graphically how pulling
$\mathbf{p}^{(\epsilon)}$ as close as possible to $\mathbf{p}^{(0)}$ tends to pull
$\mathbf{p}^{(\infty)}(\theta)$ close to $\mathbf{p}^{(0)}$ as well.
}
\label{fig:KL}
\end{figure}
Contrastive divergence \cite{Welling:2002p3,Hinton02} is a variation on steepest gradient descent of the maximum (log) likelihood (ML) objective function.  Rather than integrating over the full model distribution, CD approximates the partition function term in the gradient by averaging over the distribution obtained after taking a few Markov chain Monte Carlo (MCMC) steps away from the data distribution\footnote{
The update rule for gradient descent of the negative log likelihood, or maximum likelihood objective function, is
\begin{equation*}\label{eq:contrastive_divergence}
	\Delta \theta
\propto
	\frac
		{\partial \left[ \sum_i  p_i^{(0)} \log p_i^{(\infty)}\left(\theta\right) \right] }
		{\partial \theta}
=
	-\sum_i
		\frac{\partial E_i\left(\theta\right)}{\partial \theta}\, p_i^{(0)}
	+\sum_i
		\frac{\partial E_i\left(\theta\right)}{\partial \theta}\, p_i^{(\infty)}\left(\theta\right),
\end{equation*}
where $\mathbf{p}^{(0)}$ and $\mathbf{p}^{(\infty)}\left( \theta \right)$ represent the data distribution
and model distribution, respectively, $\mathbf{E}\left(\theta\right)$ is the energy function associated with the model distribution and $i$ indexes the states of the system (see Section~\ref{sec:distributions}).  The second term in this gradient can be extremely difficult to compute (costing in general an amount of time exponential in the dimensionality of the system).  Under contrastive divergence $p_i^{(\infty)}\left(\theta\right)$ is replaced by samples only a few Monte Carlo steps away from the data.
}.  Qualitatively, one can imagine that the data distribution is contrasted against a distribution which has evolved only a small distance towards the model distribution, whereas it would contrasted against the true model distribution in traditional MCMC approaches.  Although CD is not guaranteed to converge to the right answer, or even to a fixed point, it has proven to be an effective and fast heuristic for parameter estimation \cite{MacKay:2001p8372,Yuille04}.

Score matching, developed by Aapo Hyv\"arinen \cite{Hyvarinen05}, is a method that learns parameters in a probabilistic model using only derivatives of the energy function evaluated over the data distribution (see Equation \eqref{eq:score matching}).  This sidesteps the need to explicitly sample or integrate over the model distribution. In score matching one minimizes the expected square distance of the score function with respect to spatial coordinates given by the data distribution from the similar score function given by the model distribution.  A number of connections have been made between score matching and other learning techniques \cite{Hyvarinen:2007p5984, sohldickstein, Movellan:2008p7643, siwei2009}.

Minimum velocity learning is an approach recently proposed by
Movellan \cite{Movellan:2008p7643} that recasts a number of the ideas behind CD, treating the
minimization of the initial dynamics away from the data distribution as the goal
itself rather than a surrogate for it.  Movellan's proposal is that
rather than directly minimize the difference between the data and the model,
one introduces system dynamics that have the model as their equilibrium distribution,
and minimizes the initial flow of probability away from the data under those
dynamics. If the model looks exactly like the data there
will be no flow of probability, and if model and data are similar the flow of probability will tend to be minimal.  Movellan applies this intuition to the specific case of distributions over continuous state spaces evolving via diffusion dynamics, and recovers the score matching objective function.  The velocity in minimum velocity learning is the difference in average drift velocities between particles diffusing under the model distribution and particles diffusing under the data distribution.

Here we propose a framework called minimum probability flow learning (MPF), applicable to {\it{any}} parametric model, of which minimum velocity, SM and certain forms of CD are all special cases, and which is in many situations more powerful than any of these algorithms.  We demonstrate that learning under this framework is effective and
fast in a number of cases: Ising models  \cite{RevModPhys.39.883,Ackley85}, deep belief networks \cite{Hinton2006},
 and the product of Student-t tests model for natural scenes \cite{Welling:2003p7517}.

\section{Minimum probability flow}

Our goal is to find the parameters that cause a probabilistic model to best
agree with a set $\mathcal{D}$ of (assumed iid) observations of the state of a system.  We
will do this by proposing dynamics that guarantee the transformation of the data
distribution into the model distribution, and then minimizing the KL divergence 
which results from running those dynamics for a short time $\epsilon$ (see Figure~\ref{fig:KL}).

\subsection{Distributions}\label{sec:distributions}

The data distribution is represented by a vector $\mathbf{p}^{(0)}$, with
$p^{(0)}_i$ the fraction of the observations $\mathcal{D}$ in state $i$.  The superscript $(0)$ represents time $t=0$ under the system
dynamics (which will be described in more detail in Section~\ref{sec:dynamics}).  For example, in 
 a two variable binary system, $\mathbf{p}^{(0)}$ would have
four entries representing the fraction of the data in states $00$, $01$,
$10$ and $11$ (Figure \ref{fig:dynamics}).

Our goal is to find the parameters $\theta$ that cause a model distribution
$\mathbf{p}^{(\infty)}\left( \theta \right)$ to best match the data
distribution $\mathbf{p}^{(0)}$. The superscript
$(\infty)$ on the model distribution indicates that this is the equilibrium distribution reached after
running the dynamics for infinite time. Without loss of generality, we assume the
model distribution is of the form
\begin{eqnarray}
\label{eqn:p infinity}
p^{(\infty)}_i\left( \theta \right) = \frac
			{\exp \left( -E_i\left( \theta \right) \right) }
			{Z\left(\theta\right)}\label{p_infinity}
			,
\end{eqnarray}
where $\mathbf{E}\left( \theta \right)$ is referred to as the energy function,
and the normalizing factor $Z\left(\theta\right)$ is the partition
function, 
\begin{eqnarray}
Z\left(\theta\right) = \sum_i \exp \left( -E_i\left( \theta \right) \right)
\end{eqnarray}
(here we have set the ``temperature" of the system to 1).

\subsection{Dynamics}\label{sec:dynamics}
\begin{figure}
\center{
\framebox[0.6\linewidth]{
\parbox[c]{0.6\linewidth}{
\center{
\includegraphics[width=0.9\linewidth]{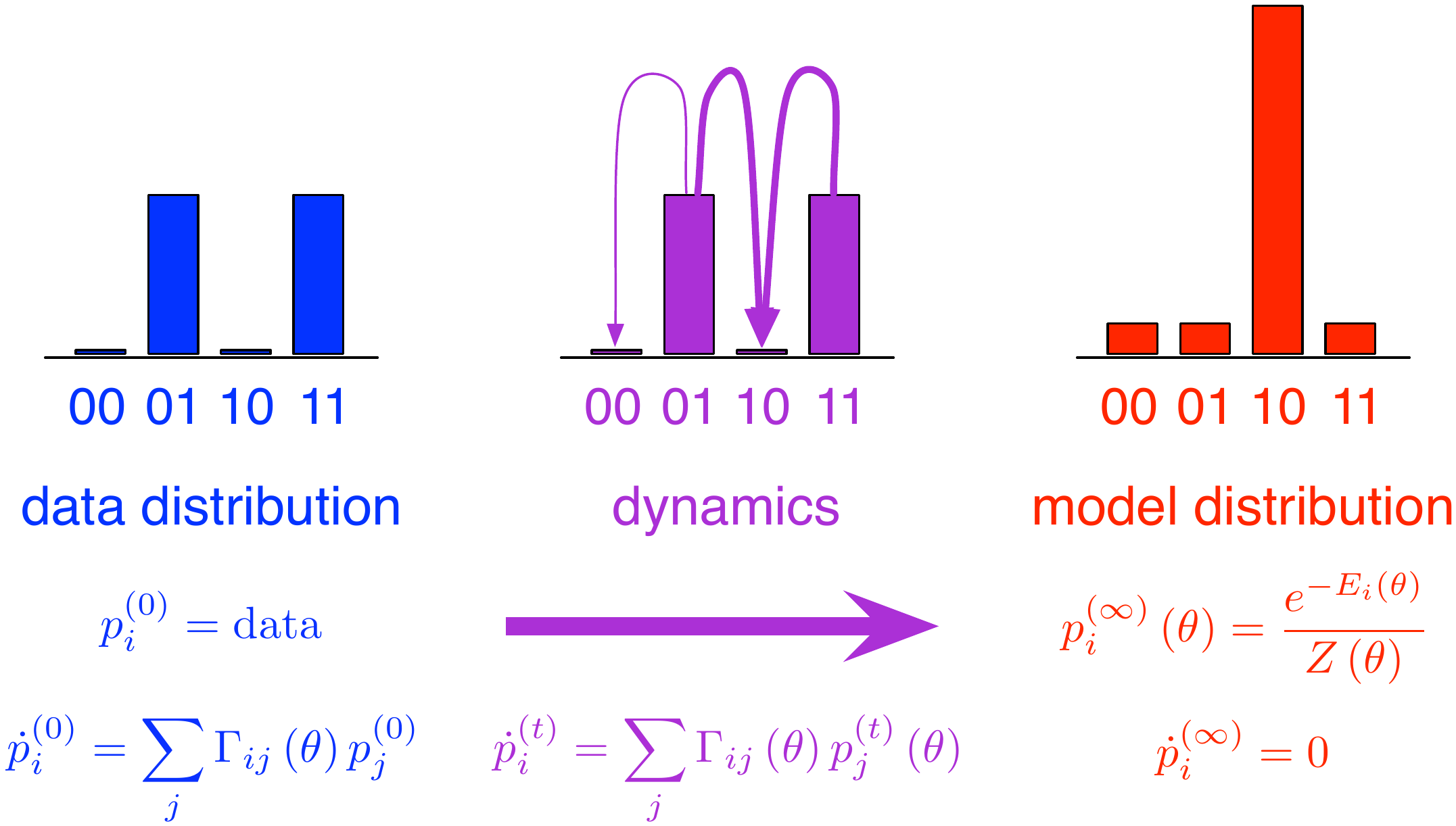}
}
}
}
}
\caption{
Dynamics of minimum probability flow learning. Model dynamics represented by
the probability flow matrix $\mathbf{\Gamma}$ ({\it middle}) determine how probability flows from
the empirical histogram of the sample data points ({\it left}) to the equilibrium
distribution of the model ({\it right}) after a sufficiently long time. In
this example there are only four possible states for the system, which consists
of a pair of binary variables, and the particular model parameters favor state
$10$ whereas the data falls on other states. 
}
\label{fig:dynamics}
\end{figure}

Most Monte-Carlo algorithms rely on two core concepts from statistical physics, 
the first being conservation of probability as enforced by the master equation for the evolution of a distribution $\mathbf p^{(t)}$ with time \cite{Pathria:1972p5861}:
\begin{equation}\label{eq:mastereqn}
\dot{p}_i^{(t)} = \sum_{j\neq i} \Gamma_{ij}(\theta)\,p_j^{(t)} - \sum_{j\neq i} \Gamma_{ji}(\theta)\, p_i^{(t)},
\end{equation}
where $\dot{p}_i^{(t)}$ is the time derivative of $p_i^{(t)}$.  Transition rates $\Gamma_{ij}(\theta)$, where $i \neq j$, give the rate at which probability flows from a
state $j$ into a state $i$.
The first term of Equation~\eqref{eq:mastereqn} captures the flow of probability out of other states $j$ into the state $i$, and the second
captures flow out of $i$ into other states $j$.
The dependence on $\theta$ results from the requirement that the chosen dynamics cause $\mathbf{p}^{(t)}$ to flow to the equilibrium distribution $\mathbf{p}^{(\infty)}(\theta)$. For readability, explicit dependence on $\theta$ will be dropped except where necessary. 
If we choose to set the diagonal elements of $\mathbf{\Gamma}$ to obey $\Gamma_{ii} = -\sum_{j\neq i}\Gamma_{ji}$, then we can write the
dynamics as
\begin{equation}\label{eq:dynamics}
\dot{\mathbf{p}}^{(t)} = \mathbf{\Gamma}\mathbf{p}^{(t)}
\end{equation}
(see Figure~\ref{fig:dynamics}).  The unique solution for $\mathbf{p}^{(t)}$ is\footnote{ The form chosen for
$\mathbf{\Gamma}$ in Equation~\eqref{eq:dynamics}, coupled with the satisfaction of detailed balance and ergodicity introduced in section \ref{sec:detailed balance}, 
 guarantees that there
is a unique eigenvector $\mathbf{p}^{(\infty)}$ of $\mathbf{\Gamma}$ with
eigenvalue zero, and that all other eigenvalues of $\mathbf{\Gamma}$ have negative real parts.}
\begin{equation}
\mathbf{p}^{(t)} = \exp\left(\mathbf{\Gamma}t\right)\mathbf{p}^{(0)}.
\end{equation}

\subsection{Detailed Balance}
\label{sec:detailed balance}

The second core concept is detailed balance,
\begin{equation}
\label{eq:detailed_balance}
\Gamma_{ji}\ p^{(\infty)}_i\left(\theta\right) = \Gamma_{ij}\ p^{(\infty)}_j\left(\theta\right)
,
\end{equation}
which states that at equilibrium the probability flow from state $i$ into state $j$ equals the probability flow from $j$ into $i$.  When satisfied, detailed balance guarantees that the distribution $\mathbf p^{(\infty)}\left(\theta\right)$ is a fixed point of the dynamics.  Sampling in most Monte Carlo methods is performed by choosing $\mathbf \Gamma$ consistent with Equation \ref{eq:detailed_balance} (and the added requirement of ergodicity), then stochastically running the dynamics of Equation \ref{eq:mastereqn}.
Note that there is no need to restrict the dynamics defined by $\mathbf{\Gamma}$ to those of any real physical process, such as diffusion. 

Equation \ref{eq:detailed_balance} can be written in terms of the model's energy function $\mathbf E\left( \theta \right)$ by substituting in Equation~\ref{eqn:p infinity} for $\mathbf p^{(\infty)}\left(\theta\right)$:
\begin{equation}
	{\Gamma_{ji}}
	\exp \left( -E_i\left( \theta \right) \right)
=
	{\Gamma_{ij}}
	\exp \left( -E_j\left( \theta \right) \right)
.
\end{equation}
$\mathbf{\Gamma}$ is underconstrained by the above equation. Motivated by symmetry and aesthetics, we choose as the form for the non-diagonal entries in $\mathbf{\Gamma}$
\begin{eqnarray}
\label{eqn:gamma symmetric}
 \ \ \ \ \ \ \ \ \ \ \ \ \ \ \ \ \ \ \Gamma_{ij} & = &
	g_{ij} \exp \left[ \frac{1}{2} \left( E_j\left( \theta \right)-E_i\left(  \theta \right) \right)\right]\ \ \ \ \ \ \ \ \ \ \ \ \ \ \ \ \ \ \ \ \ \ \ \left(i \neq j\right) 
,
\end{eqnarray} 
where the connectivity function
\begin{eqnarray}
\label{eqn:gamma symmetric}
\ \ \ \ \ \ \ \ \ \
	g_{ij} = g_{ji} = & 
	\left\{\begin{array}{ccc}
		0 &  & \mathrm{unconnected\ states} \\
		1 &  & \mathrm{connected\ states}
	\end{array}\right.
\ \ \ \ \ \ \ \ \ \ \ \ \ \ \ \ \ \ \ \ \ \ \ \left(i \neq j\right)
\end{eqnarray} 
determines which states are allowed to directly exchange probability with each other\footnote{The non-zero $\mathbf \Gamma$ may also be sampled from a proposal distribution rather than set via a deterministic scheme, in which case $g_{ij}$ takes on the role of proposal distribution, and a factor of 
$\left( \frac
	{g_{ji}}
	{g_{ij}}
\right)^{\frac{1}{2}}$
enters into $\Gamma_{ij}$.}.  $g_{ij}$ can be set such that $\mathbf \Gamma$ is {\em extremely} sparse (see Section~\ref{sec:tractability}).  Theoretically, to guarantee convergence to the model distribution, the non-zero elements of $\mathbf{\Gamma}$ must be chosen such that, given sufficient time, probability can flow between any pair of states.

\subsection{Objective Function}

Maximum likelihood parameter estimation involves maximizing the likelihood of some observations $\mathcal{D}$ under a model, or equivalently minimizing the KL divergence between the data distribution $\mathbf p^{(0)}$ and model distribution $\mathbf p^{(\infty)}$,
\begin{eqnarray}
\hat{\theta}_{\mathrm{ML}} & = & \arg \min_\theta D_{KL}\left( 
\mathbf{p^{(0)}} ||\mathbf{p^{(\infty)}}
\left(\theta\right)\right)
\end{eqnarray}
Rather than running the dynamics for infinite time, we propose to minimize the KL divergence after running the dynamics for an infinitesimal time $\epsilon$,
\begin{eqnarray}
\hat{\theta}_{\mathrm{MPF}} & = & \arg \min_\theta K\left( \theta \right) \\
K\left( \theta \right) & = & 
	D_{KL}\left( 
	\mathbf{p^{(0)}} ||\mathbf{p^{(\epsilon)}}
	\left(\theta\right)\right)
.
\end{eqnarray}
For small $\epsilon$, $D_{KL}\left( 
	\mathbf{p^{(0)}} ||\mathbf{p^{(\epsilon)}}
	\left(\theta\right)\right)$ can be approximated by a first order Taylor expansion,
\begin{eqnarray}
K\left( \theta \right) & \approx & D_{KL}\left( 
	\mathbf{p^{(0)}} ||\mathbf{p^{(t)}}
	\left(\theta\right)\right)\Big |_{t=0} 
	+ \epsilon \frac
	{\partial D_{KL}\left( 
	\mathbf{p^{(0)}} ||\mathbf{p^{(t)}}
	\left(\theta\right)\right)}
	{\partial t}\Big |_{t=0}
.
\end{eqnarray}
Further algebra (see Appendix \ref{app:KL}) reduces $K\left( \theta \right)$ to a measure of the flow of probability, at time $t = 0$ under the dynamics, out of data states $\mathcal{D}$ into non-data states,
\begin{eqnarray}
\label{eq:Kfinal}
K\left( \theta \right) & = & \frac{\epsilon}{|\mathcal{D}|}\sum_{i\notin \mathrm{\mathcal{D}}}\sum_{j\in \mathrm{\mathcal{D}}} \Gamma_{ij}
= \frac{\epsilon}{|\mathcal{D}|} \sum_{j\in \mathrm{\mathcal{D}}
 } \sum_{
		 i\notin \mathrm{\mathcal{D}} 
 } 
g_{ij} \exp \left[ \frac{1}{2} \left( E_j\left( \theta \right)-E_i\left(  \theta \right) \right)\right] 
\end{eqnarray}
with gradient\footnote{The contrastive divergence update rule can be written in the form
\begin{eqnarray}
\label{eq:CDstep}
\Delta \theta_{CD} \propto 
	-\sum_{j\in \mathrm{\mathcal{D}}
 } \sum_{
		 i\notin \mathrm{\mathcal{D}} 
 } 
\left[ \pd{E_j\left( \theta \right)}{\theta}-\pd{E_i\left(  \theta \right)}{\theta} \right] \mathrm{[probability\ of\ MCMC\ step\ from\ j\rightarrow i]}
\end{eqnarray}
with obvious similarities to the MPF learning gradient.  Thus steepest gradient descent under MPF resembles CD updates, but with the sampling/rejection step replaced by a weighting factor.  Note that this difference in form provides MPF with an objective function, and guarantees a unique global minimum when model and data distributions agree.}
\begin{eqnarray}
\label{eq:Kgrad}
&& \pd{K\left( \theta \right) }{\theta}
 = \frac{\epsilon}{|\mathcal{D}|} \sum_{j\in \mathrm{\mathcal{D}}
 } \sum_{
		 i\notin \mathrm{\mathcal{D}} 
 } 
\left[ \pd{E_j\left( \theta \right)}{\theta}-\pd{E_i\left(  \theta \right)}{\theta} \right]
g_{ij} \exp \left[ \frac{1}{2} \left( E_j\left( \theta \right)-E_i\left(  \theta \right) \right)\right] 
 ,
\end{eqnarray}
where $|\mathcal{D}|$ is the number of observed data points.
Note that Equations~\eqref{eq:Kfinal} and~\eqref{eq:Kgrad} do not depend on the partition function $Z\left( \theta \right)$ or its derivatives.

$K\left( \theta \right)$ is uniquely zero when $\mathbf{p}^{(0)}$ and $\mathbf{p}^{(\infty)}\left( \theta \right)$ are exactly equal (although in general $K\left( \theta \right)$ provides a lower bound rather than an upper bound on 
$D_{KL}\left( 
\mathbf{p^{(0)}} ||\mathbf{p^{(\infty)}}
\left(\theta\right)\right)$
).  In addition, $K\left( \theta \right)$ is convex for all models $\mathbf p^{(\infty)}\left( \theta \right)$ in the exponential family - that is, models whose energy functions ${\mathbf E}\left( \theta \right)$ are linear in their parameters $\theta$ \cite{macke} (see Appendix \ref{app:convex}).

\subsection{Tractability}
\label{sec:tractability}
The vector $\mathbf{p}^{(0)}$ is typically huge, as is 
$\mathbf{\Gamma}$ ({\it{e.g.}}, $2^N$ and $2^N \times 2^N$, respectively, for an $N$-bit binary
system).  Na\"ively, this would seem to prohibit evaluation and minimization of
the objective function.  Fortunately, all the elements in $\mathbf{p}^{(0)}$
not corresponding to observations are zero.  
This allows us to ignore all those $\Gamma_{ij}$ for which no data point exists at state $j$.
Additionally, there is a great deal
of flexibility as far as which elements of $\mathbf g$, and thus $\mathbf{\Gamma}$, can be set to zero.
By populating $\mathbf{\Gamma}$ so as to connect each state to a small fixed number of additional states,
the cost of the algorithm in both memory and time is $\mathcal{O}(|\mathcal{D}|)$, which does not depend
on the number of system states, only on the number of observed data points.

\subsection{Continuous Systems}\label{sec:cont_systems}

Although we have motivated this technique using systems with a large, but
finite, number of states, it generalizes in a straightforward manner to
continuous systems.  The flow matrix $\mathbf{\Gamma}$ and distribution vectors $\mathbf{p}^{(t)}$ transition from being very large to being infinite in size.  $\mathbf{\Gamma}$ can still be chosen to connect each state to a small, finite, number of additional states however, and only outgoing probability flow from states with data contributes to the objective function, so the cost of learning remains largely unchanged.

In addition, for a particular pattern of connectivity in $\mathbf{\Gamma}$ this objective function, like
Movellan's \cite{Movellan:2008p7643}, reduces to score matching \cite{Hyvarinen05}.
Taking the limit of connections between all states within a small distance $\epsilon_x$ of each other, and then Taylor expanding in $\epsilon_x$, one can show that, up to an overall constant and scaling factor
\begin{equation}
\label{eq:score matching}
K = K_{\mathrm{SM}} = \ \sum_{i\in\mathcal{D}} \left[ \frac{1}{2}\nabla E(x_i)\cdot \nabla E(x_i) - \nabla^2 E(x_i) \right ].
\end{equation}
This reproduces the link discovered by Movellan \cite{Movellan:2008p7643} between diffusion dynamics over continuous spaces and score matching.

\section{Experimental Results}
Matlab code implementing minimum probability flow learning for each of the following cases is available upon request.  A public toolkit is under construction.

All minimization was performed using Mark Schmidt's remarkably effective minFunc \cite{schmidt}.

\subsection{Ising model}
\begin{figure}
\center{
\parbox[c]{\textwidth}{
\center{
\includegraphics[width= 0.4 \linewidth]{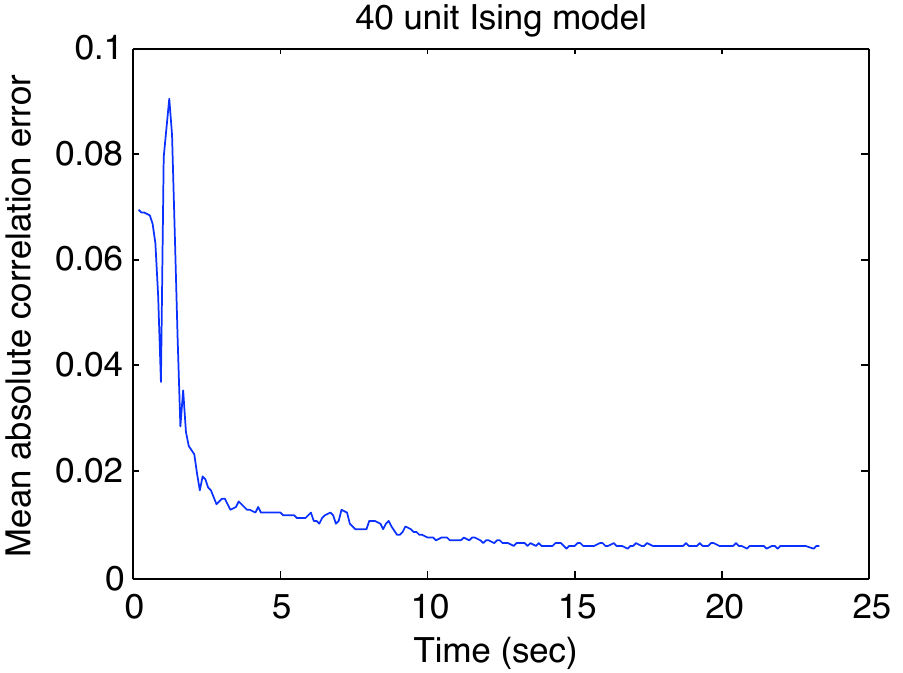}
\ \ \ \ \ \ \ \ \ \ \ \ 
\includegraphics[width= 0.4 \linewidth]{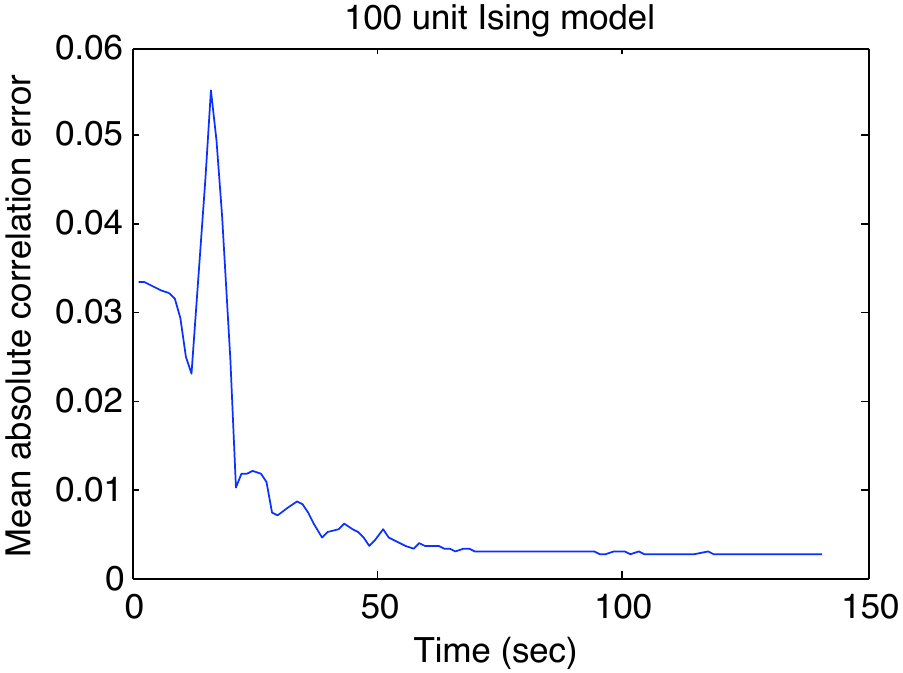}
}
}
}
\caption{
A demonstration of rapid fitting of the Ising model by minimum probability flow learning.  The mean absolute error in the learned model's correlation matrix is shown as a functions of learning time for 40 and 100 unit fully connected Ising models.  Convergence is reached in about $15$ seconds for $20,000$ samples from the 40 unit model \emph{(left)} and in about 1 minute for $100,000$ samples from the 100 unit model \emph{(right)}.  Details of the 100 unit model can be seen in Figure~\ref{fig:ising100}.
}
\label{fig:ising time}
\end{figure}
\begin{figure}
\center{
\parbox[c]{\textwidth}{
\center{
\includegraphics[width= 0.9\linewidth]{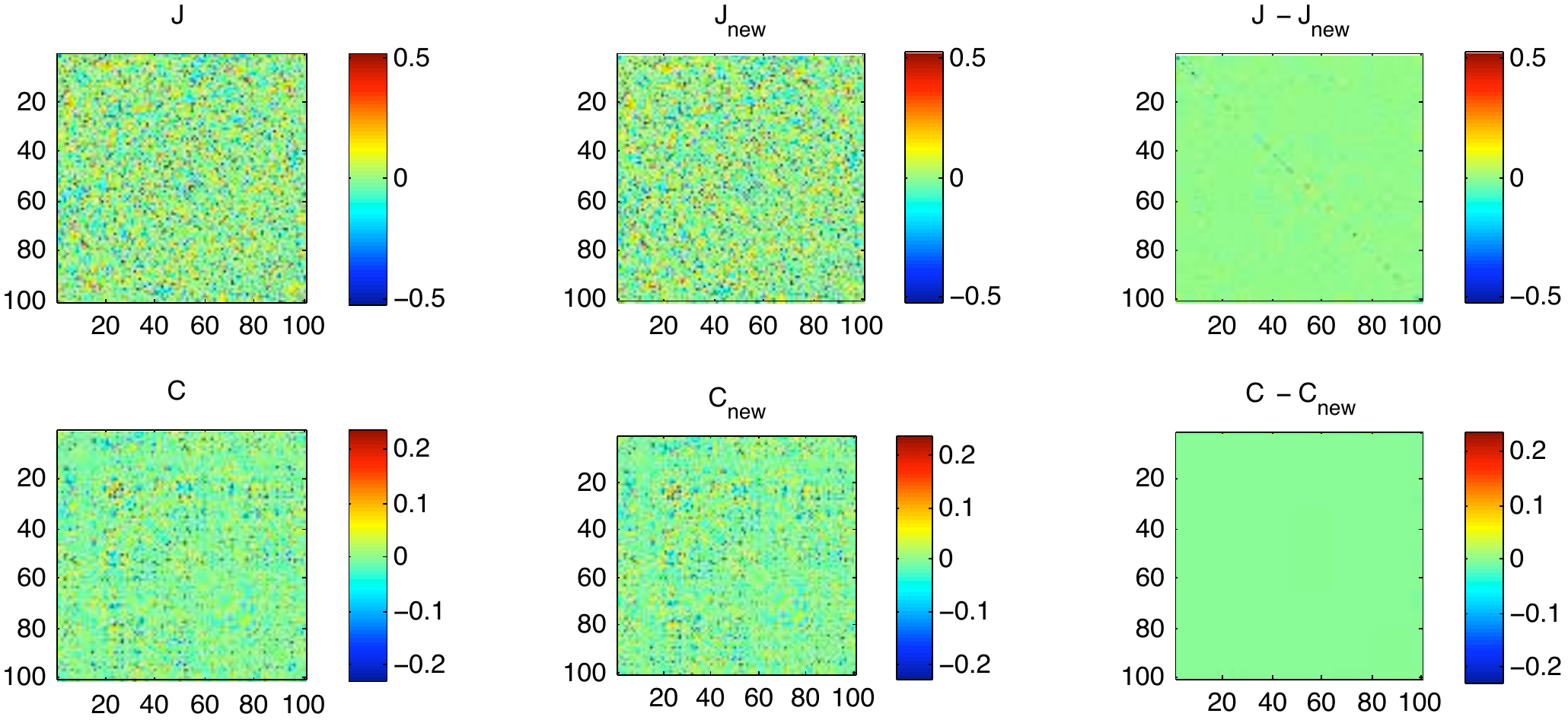}
}
}
}
\caption{
An example 100 unit Ising model fit using minimum probability flow learning. \emph{(left)} Randomly chosen Gaussian coupling matrix $J$ (top) with variance $0.04$ and associated correlation matrix $C$ (bottom) for a 100 unit, fully-connected Ising model.  The diagonal has been removed from the correlation matrix $C$ for increased visibility. \emph{(center)} The recovered coupling and correlation matrices after minimum probability flow learning on $100,000$ samples from the model in the left panels.  \emph{(right)} The error in recovery of the coupling and correlation matrices.
}
\label{fig:ising100}
\end{figure}

The Ising model has a long and storied history in physics \cite{RevModPhys.39.883} and machine learning \cite{Ackley85} and it has recently been found to be a surprisingly useful model for networks of neurons in the retina \cite{Schneidman_Nature_2006,Shlens_JN_2006}.  The ability to fit Ising models to the activity of large groups of simultaneously recorded neurons is of current interest given the increasing number of these types of data sets from the retina, cortex and other brain structures.

We fit an Ising model (fully visible Boltzmann machine) of the form
\begin{equation}
p^{(\infty)}(\mathbf{x};\mathbf{J}) = \frac{1}{Z(\mathbf{J})}\exp\left[ -\sum_{i,j} J_{ij} x_i x_j \right]
\end{equation}
to a set of $N$ $d$-element iid data samples $\left\{x^{(i)} | i = 1...N\right\}$ generated via Gibbs sampling from an Ising model as described below, where each of the $d$ elements of $\mathbf{x}$ is either 0 or 1.
Because each $x_i\in\set{0,1}$, $x_i^2=x_i$, we can write the energy function as
\begin{equation}
E(\mathbf{x};\mathbf{J}) = \sum_{i, j\neq i} J_{ij} x_i x_j + \sum_{i} J_{ii} x_i.
\end{equation}

The probability flow matrix $\mathbf{\Gamma}$ has $2^N \times 2^N$ elements, but for learning we populate it extremely sparsely,
setting
\begin{eqnarray}
\label{eqn:gamma symmetric}
	g_{ij} = g_{ji} = & 
	\left\{\begin{array}{ccc}
		1 &  & \mathrm{states\ }i\mathrm{\ and\ }j\mathrm{\ differ\ by\ single\ bit\ flip} \\
		0 &  & \mathrm{otherwise}
	\end{array}\right.
.\label{eq:gij_ising}
\end{eqnarray} 

Figure~\ref{fig:ising time} shows the average error in predicted
correlations as a function of learning time for $20,000$ samples from a 40 unit, fully connected Ising model.  The $J_{ij}$ used were graciously provided by Broderick and coauthors, and were identical to those used for synthetic data generation in the 2008 paper ``Faster solutions of the inverse pairwise Ising problem" \cite{Broderick:2007p2761}.  Training was performed on $20,000$ samples so as to match the number of samples used in section III.A. of Broderick et al.  Note that given sufficient samples, the minimum probability flow algorithm would converge exactly to the right answer, as learning in the Ising model is convex (see Appendix \ref{app:convex}), and has its global minimum at the true solution.  On an 8 core 2.33 GHz Intel Xeon,
the learning converges in about $15$ seconds. Broderick et al. perform a similar learning task on a 100-CPU grid computing
cluster, with a convergence time of approximately $200$ seconds.

Similar learning was performed for $100,000$ samples from 
a 100 unit, fully connected Ising model.  A coupling matrix was chosen with elements randomly drawn from a Gaussian with mean $0$ and variance $0.04$.
 Using the minimum probability flow
learning technique, learning took approximately 1 minute, compared to roughly 12 hours for a 100 unit (nearest neighbor coupling only) model of retinal data \cite{Shlens:2009p4887} (personal communication, J. Shlens).
Figure~\ref{fig:ising100} demonstrates the recovery of the coupling and correlation matrices for our fully connected Ising model, while Figure~\ref{fig:ising time} shows the time course for learning.

\subsection{Deep Belief Network}
\begin{figure}
\center{
\parbox[c]{\textwidth}{
\center{
\includegraphics[width=0.15 \linewidth]{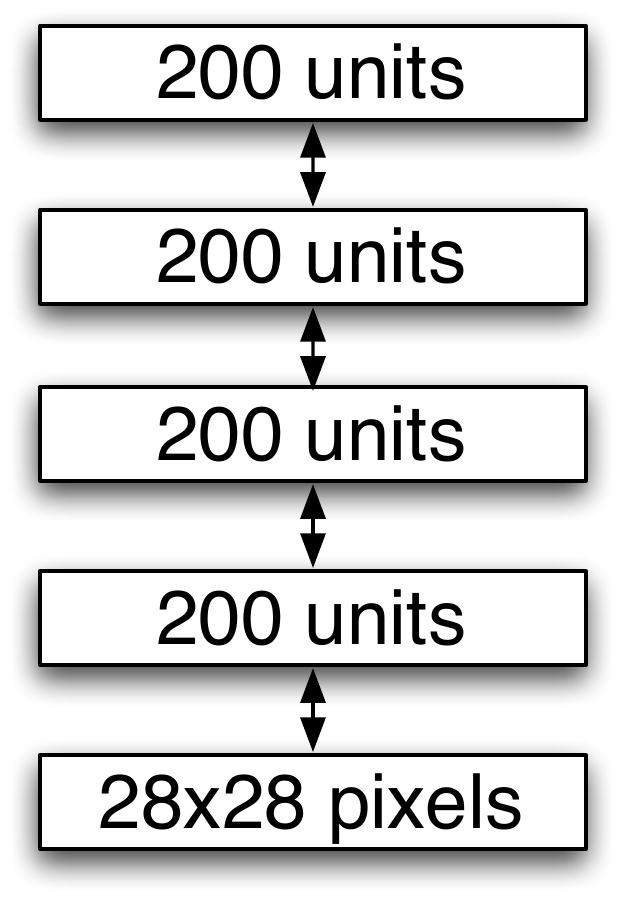}
\ \ \ \ \ \ \ \ \ \ \ \ \ \ \ \ \ \ 
\includegraphics[width=0.22 \linewidth]{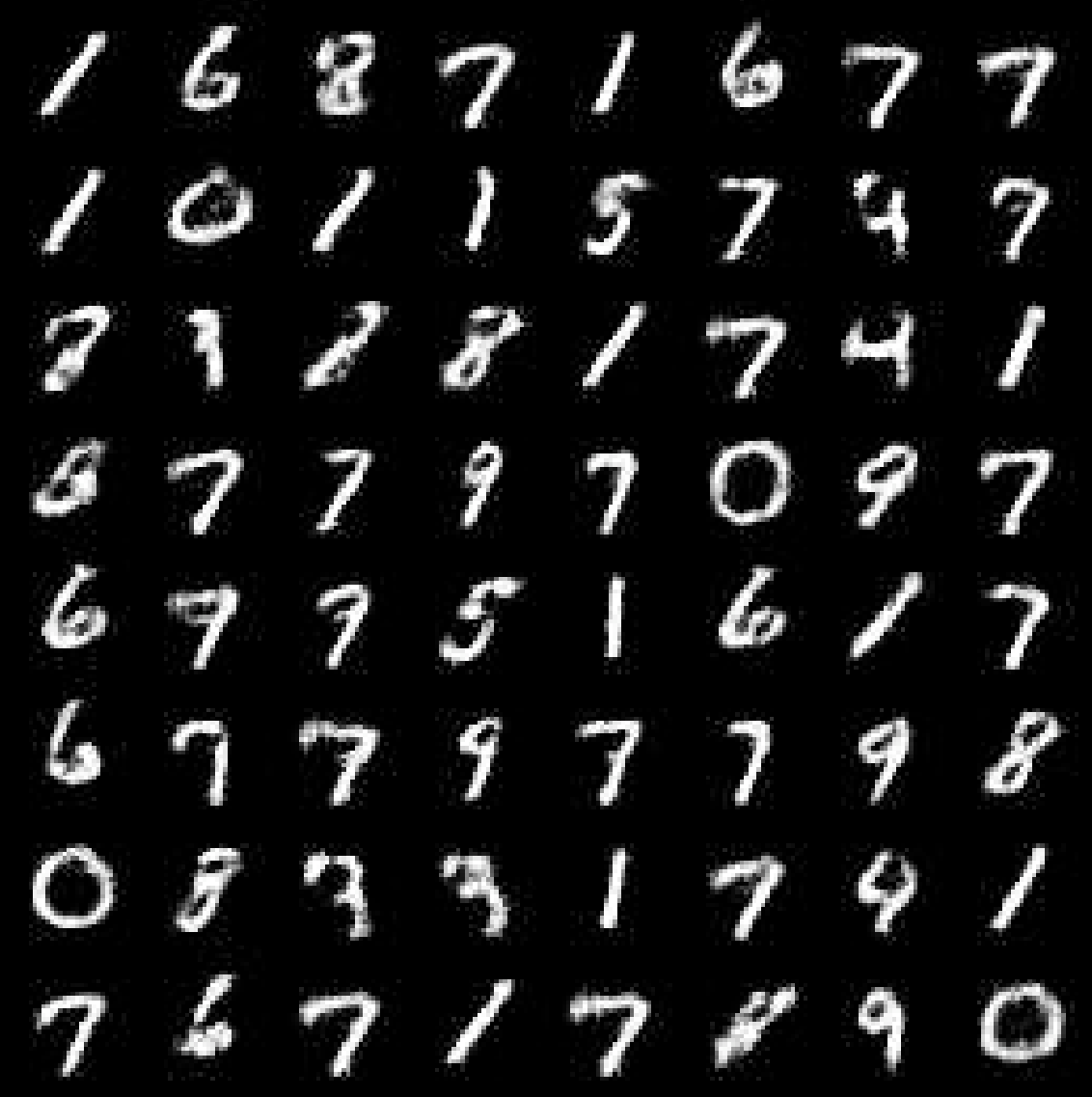}
\ \ \ \ \ \ \ \ \ \ \ \ \ \ \ \ \ \ 
\includegraphics[width=0.2214 \linewidth]{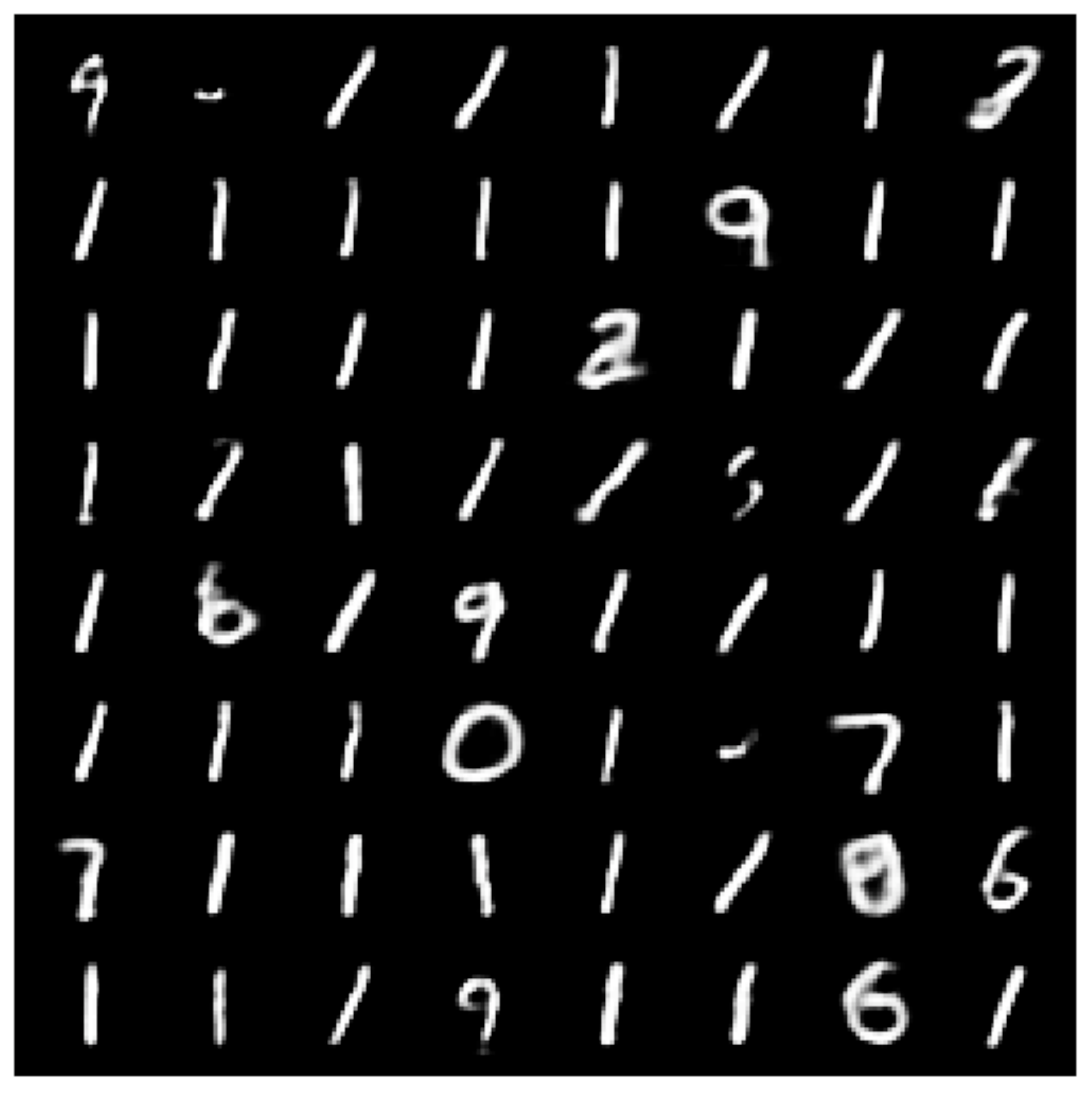}
}
} 
}
\caption{
A deep belief network trained using minimum probability flow learning and contrastive divergence. \emph{(left)} A four layer deep belief network was trained on the MNIST postal hand written digits dataset. \emph{(center)} Confabulations after training via minimum probability flow learning.  A reasonable probabilistic model for handwritten digits has been learned.  \emph{(right)}  Confabulations after training via single step CD.  Note the uneven distribution of digit occurrences.
}
\label{fig:dbn}
\end{figure}
As a demonstration of learning on a more complex discrete valued model, we trained a 4 layer deep belief network (DBN) \cite{Hinton2006} on MNIST handwritten digits.  A DBN consists of stacked restricted Boltzmann machines (RBMs), such that the hidden layer of one RBM forms the visible layer of the next.  Each RBM has the form:
\begin{eqnarray}
p^{(\infty)}(\mathbf{x}_{\mathrm{vis}}, \mathbf{x}_{\mathrm{hid}};\mathbf{W}) & = & \frac{1}{Z(\mathbf{W})}\exp\left[ -\sum_{i,j} W_{ij} x_{\mathrm{vis},i} x_{\mathrm{hid},j} \right]
,
 \\
p^{(\infty)}(\mathbf{x}_{\mathrm{vis}};\mathbf{W}) & = & \frac{1}{Z(\mathbf{W})}\exp\left[ 
\sum_{j} \log
	\left(
	1 +
	\exp\left[
		-\sum_i W_{ij} x_{\mathrm{vis},i}
	\right]
	\right)
 \right]
 .
\end{eqnarray}
Note that sampling-free application of the minimum probability flow algorithm requires analytically marginalizing over the hidden units.  RBMs were trained in sequence, starting at the bottom layer, on 10,000 samples from the MNIST postal hand written digits data set. As in the Ising case, the probability flow matrix $\mathbf{\Gamma}$ was populated so as to connect every state to all states which differed by only a single bit flip (see Equation \ref{eq:gij_ising}).  Training was performed by both minimum probability flow and single step CD to allow a simple comparison of the two techniques (note that CD turns into full ML learning as the number of steps is increased, and that the quality of the CD answer can thus be improved at the cost of computational time by using many-step CD).

Confabulations were performed by Gibbs sampling from the top layer RBM, then propagating each sample back down to the pixel layer by way of the conditional distribution $p^{(\infty)}(\mathbf{x}_{\mathrm{vis}} | \mathbf{x}_{\mathrm{hid}};\mathbf{W}^k)$ for each of the intermediary RBMs, where $k$ indexes the layer in the stack.  As shown in Figure~\ref{fig:dbn}, minimum probability flow learned a good model of handwritten digits.


\subsection{Product of Student-t Model}
\begin{figure}
\center{
\parbox[c]{\textwidth}{
\center{
\includegraphics[width= 0.4\linewidth]{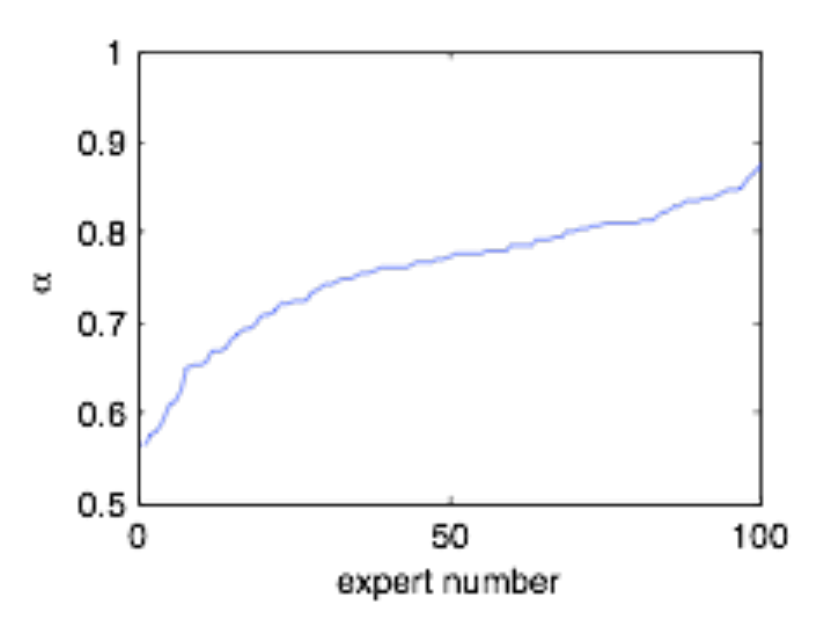}
\ \ \ \ \ \ \ \ \ \ \ \ \ \ \ \ \ \ \ \ \ 
\includegraphics[width= 0.3\linewidth]{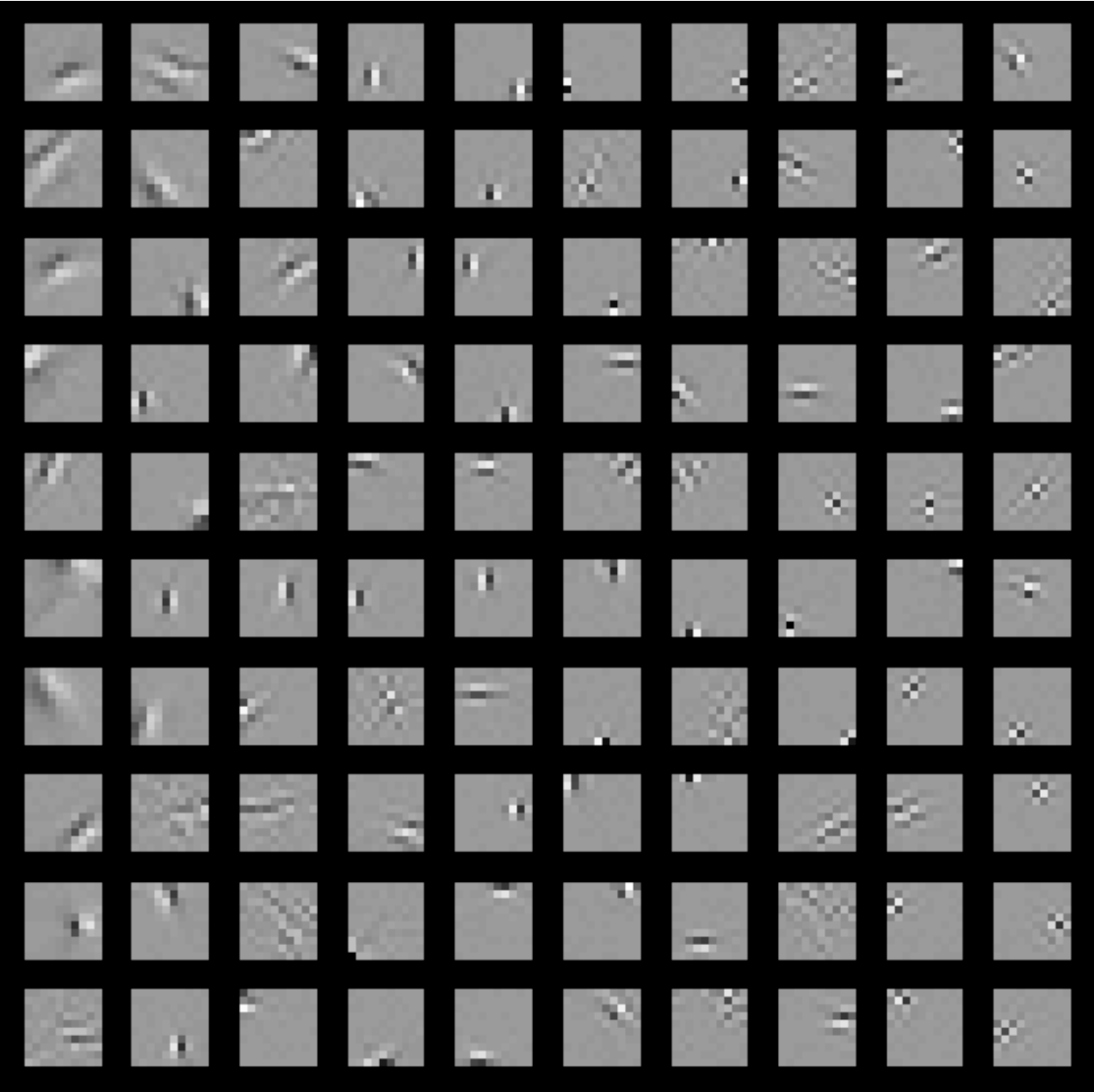}
}
}
}
\caption{
A continuous state space model fit using minimum probability flow learning. The sparsity parameter $\mathbf \alpha$ \emph{(left)} and receptive fields $\mathbf J$ \emph{(right)} are shown for a product of Student-t model trained on natural images.  Both plots are ordered by sparsity.
}
\label{fig:student-t}
\end{figure}
As a demonstration of parameter estimation in continuous state space, non-analytically normalizable, probabilistic models, we trained a product of Student-t model \cite{Welling:2003p7517} with 100 receptive fields $\mathbf J$ on  $100,000$ $10 \times 10$ natural image patches.  The  product of Student-t model has the form
\begin{eqnarray}
p^{(\infty)}\left(\mathbf x; \mathbf J, \mathbf\alpha\right)
\propto
e^{
-\sum_i \alpha_i \log \left[
1 + \left(
\mathbf J_{i} \mathbf x
\right)^2
\right]
}
,
\end{eqnarray}
where the $\mathbf \alpha$ parameter controls the sparsity of the distribution.  $g_{ij}$ was chosen such that each state $\mathbf x$ was connected to 2 states $\tilde{\mathbf x}$ chosen by adding Gaussian noise to x and rescaling to maintain patch variance --- that is $\tilde{\mathbf x} = \left(\mathbf x + \tilde{\mathbf n}\right)\frac{||\mathbf x||_2}{||\mathbf x+\tilde{\mathbf n}||_2}$ for Gaussian noise vector $\tilde{\mathbf n} \sim N\left( 0, 0.1 \right)$. 
Figure \ref{fig:student-t} shows that this model learns oriented edges, as expected.

\section{Summary}
We have presented a novel framework for efficient learning in the context of any parametric model.  This method was inspired by the minimum velocity approach developed by Movellan, and it reduces to that technique as well as to score matching and some forms of contrastive divergence under suitable choices for the dynamics and state space.  By decoupling the dynamics from any specific physical process, such as diffusion, and focusing on the initial flow of probability from the data to a subset of other states chosen in part for their utility and convenience, we have arrived at a framework that is not only more general than previous approaches, but also potentially much more powerful. We expect that this framework will render some previously intractable models more amenable to analysis.
\subsubsection*{Acknowledgments}

We would like to thank Javier Movellan for sharing a work in progress; Tamara Broderick, Miroslav Dud\'{\i}k, Ga\v{s}per Tka\v{c}ik, Robert E. Schapire and William Bialek for use of their Ising model coupling parameters; Jonathon Shlens for useful discussion and ground truth for his Ising model convergence times; Bruno Olshausen, Anthony Bell, Christopher Hillar, Charles Cadieu, Kilian Koepsell and the rest of the Redwood Center for many useful discussions and for comments on earlier versions of the manuscript; Ashvin Vishwanath for useful discussion; and the Canadian Institute for Advanced Research - Neural Computation and Perception Program for their financial support (JSD).

\newpage

  \begin{center}
    {\bf APPENDICES}
  \end{center}

\appendix
\section{Taylor Expansion of KL Divergence}
\label{app:KL}

 \renewcommand{\theequation}{A-\arabic{equation}}
  \setcounter{equation}{0}  

\begin{eqnarray}
K\left( \theta \right) & \approx & D_{KL}\left( 
	\mathbf{p^{(0)}} ||\mathbf{p^{(t)}}
	\left(\theta\right)\right)\Big |_{t=0} 
	+ \epsilon \frac
	{\partial D_{KL}\left( 
	\mathbf{p^{(0)}} ||\mathbf{p^{(t)}}
	\left(\theta\right)\right)}
	{\partial t}\Big |_{t=0} \\
& = & 0
	+ \epsilon \frac
	{\partial D_{KL}\left( 
	\mathbf{p^{(0)}} ||\mathbf{p^{(t)}}
	\left(\theta\right)\right)}
	{\partial t}\Big |_{t=0} \\
& = & \epsilon \pd{}{t}\left.\left(\sum_{i\in \mathcal{D}} \p{i}{0}\log\frac{\p{i}{0}}{\p{i}{t}}\right)\right|_0 \\
&=& -\epsilon \sum_{i\in \mathcal{D}}\frac{\p{i}{0}}{\p{i}{0}}\left.\pd{\p{i}{t}}{t}\right|_0 \\
&=& -\epsilon \left.\sum_{i\in \mathcal{D}}\pd{\p{i}{t}}{t}\right|_0 \\
&=& -\epsilon \left.\left(\pd{}{t}\sum_{i\in \mathcal{D}}\p{i}{t}\right)\right|_0  \label{eqn:sumrate} \\
&=& -\epsilon \left.\pd{}{t}\left(1-\sum_{i\notin \mathcal{D}}\p{i}{t}\right)\right|_0 \\
&=& \epsilon \left.\sum_{i\notin \mathcal{D}}\pd{\p{i}{t}}{t}\right|_0 \\
&=& \epsilon \sum_{i\notin \mathcal{D}}\sum_{j\in \mathcal{D}}\Gamma_{ij}\p{j}{0} \\
&=& \frac{\epsilon}{|\mathcal{D}|} \sum_{i\notin \mathcal{D}}\sum_{j\in \mathcal{D}}\Gamma_{ij},
\end{eqnarray}
where we used the fact that $\sum_{i\in \mathcal{D}} p_i^{(t)} + \sum_{i\notin \mathcal{D}} p_i^{(t)} = 1$.
This implies that the rate of growth of the KL divergence at time $t=0$ equals the total initial flow of probability from states with data into states without.

\section{Convexity}
\label{app:convex}

 \renewcommand{\theequation}{B-\arabic{equation}}
  \setcounter{equation}{0}  

As observed by Macke and Gerwinn \cite{macke}, the MPF objective function is convex for models in the exponential family.

We wish to minimize
\begin{eqnarray}
K & = & 
	\sum_{i \in D} \sum_{j \in D^C} \Gamma_{ji} p_i^{(0)} .
\end{eqnarray}

$K$ has derivative
\begin{eqnarray}
\pd{K}{\theta_m} &= & \sum_{i \in D}\sum_{j \in D^c} \left( \pd{\Gamma_{ij}}{\theta_m} \right) p_i^{(0)} \\
&=& \frac{1}{2} \sum_{i \in D}\sum_{j \in D^c} \Gamma_{ij} \left(  \pd{E_j}{\theta_m} - \pd{E_i}{\theta_m}  \right) p_i^{(0)}, 
\end{eqnarray}
and Hessian
\begin{eqnarray}
\pd{^2 K}{\theta_m \partial \theta_n}
	&= &
		\frac{1}{4} \sum_{i \in D}\sum_{j \in D^c} \Gamma_{ij} \left(  \pd{E_j}{\theta_m} - \pd{E_i}{\theta_m}  \right)\left(  \pd{E_j}{\theta_n} - \pd{E_i}{\theta_n}  \right) p_i^{(0)}  \\
		& &+
		\frac{1}{2} \sum_{i \in D}\sum_{j \in D^c} \Gamma_{ij} \left(  \pd{^2 E_j}{\theta_m \partial \theta_n} - \pd{^2 E_i}{\theta_m \partial \theta_n}  \right) p_i^{(0)} . \nonumber
\end{eqnarray}
The first term is a weighted sum of outer products, with non-negative weights $\frac{1}{4} \Gamma_{ij} p_i^{(0)}$, and is thus positive semidefinite.  The second term is $0$ for models in the exponential family (those with energy functions linear in their parameters).

Parameter estimation for models in the exponential family is therefore convex using minimum probability flow learning.

\small{
\bibliographystyle{plain} 
\bibliography{nips2010}        

\begin{thebibliography}{10}

\bibitem{Ackley85}
D~H Ackley, G~E Hinton, and T~J Sejnowski.
\newblock {A learning algorithm for Boltzmann machines.}
\newblock {\em Cognitive Science}, 9(2):147--169, 1985.

\bibitem{Broderick:2007p2761}
T~Broderick, M~Dud\'{\i}k, G~Tka\v{c}ik, R~Schapire, and W~Bialek.
\newblock {Faster solutions of the inverse pairwise Ising problem}.
\newblock {\em E-print arXiv}, Jan 2007.

\bibitem{RevModPhys.39.883}
S~G Brush.
\newblock {History of the Lenz-Ising model}.
\newblock {\em Reviews of Modern Physics}, 39(4):883--893, Oct 1967.

\bibitem{Hinton02}
M~A Carreira-Perpi$\tilde{\mathrm{n}}$\'an and G~E Hinton.
\newblock {On contrastive divergence (CD) learning}.
\newblock {\em Technical report, Dept. of Computere Science, University of
  Toronto}, 2004.

\bibitem{haykin2008nnc}
S~Haykin.
\newblock {\em Neural networks and learning machines; 3rd edition}.
\newblock Prentice Hall, 2008.

\bibitem{Hinton2006}
Geoffrey~E Hinton, Simon Osindero, and Yee-Whye Teh.
\newblock A fast learning algorithm for deep belief nets.
\newblock {\em Neural Computation}, 18(7):1527--1554, Jul 2006.

\bibitem{Hyvarinen05}
A~Hyv\"arinen.
\newblock Estimation of non-normalized statistical models using score matching.
\newblock {\em Journal of Machine Learning Research}, 6:695--709, 2005.

\bibitem{Hyvarinen:2007p5984}
A~Hyv\"arinen.
\newblock Connections between score matching, contrastive divergence, and
  pseudolikelihood for continuous-valued variables.
\newblock {\em IEEE Transactions on Neural Networks}, Jan 2007.

\bibitem{Jaakkola:1997p4985}
T~Jaakkola and M~Jordan.
\newblock {A variational approach to Bayesian logistic regression models and
  their extensions}.
\newblock {\em Proceedings of the Sixth International Workshop on Artificial
  Intelligence and Statistics}, Jan 1997.

\bibitem{Kappen:1997p6}
H~Kappen and F~Rodr{\'i}guez.
\newblock {Mean field approach to learning in Boltzmann machines}.
\newblock {\em Pattern Recognition Letters}, Jan 1997.

\bibitem{siwei2009}
S~Lyu.
\newblock Interpretation and generalization of score matching.
\newblock {\em The proceedings of the 25th conference on uncerrtainty in
  artificial intelligence (UAI*90)}, 2009.

\bibitem{MacKay:2001p8372}
D~MacKay.
\newblock Failures of the one-step learning algorithm.
\newblock {\em http://www.inference.phy.cam.ac.uk/mackay/gbm.pdf}, Jan 2001.

\bibitem{macke}
J~Macke and S~Gerwinn.
\newblock Personal communication.
\newblock 2009.

\bibitem{movellan2008cdg}
J~R Movellan.
\newblock {{Contrastive divergence in Gaussian diffusions}}.
\newblock {\em Neural Computation}, 20(9):2238--2252, 2008.

\bibitem{Movellan:2008p7643}
J~R Movellan.
\newblock A minimum velocity approach to learning.
\newblock {\em unpublished draft}, Jan 2008.

\bibitem{Movellan93}
J~R Movellan and J~L McClelland.
\newblock Learning continuous probability distributions with symmetric
  diffusion networks.
\newblock {\em Cognitive Science}, 17:463--496, 1993.

\bibitem{Pathria:1972p5861}
R~Pathria.
\newblock {\em Statistical Mechanics}.
\newblock Butterworth Heinemann, Jan 1972.

\bibitem{schmidt}
M~Schmidt.
\newblock minfunc.
\newblock {\em http://www.cs.ubc.ca/~schmidtm/Software/minFunc.html}, 2005.

\bibitem{Schneidman_Nature_2006}
E~Schneidman, M~J~Berry 2nd, R~Segev, and W~Bialek.
\newblock Weak pairwise correlations imply strongly correlated network states
  in a neural population.
\newblock {\em Nature}, 440(7087):1007--12, 2006.

\bibitem{Shlens:2009p4887}
J~Shlens, G~D Field, J~L Gauthier, M~Greschner, A~Sher, A~M Litke, and E~J
  Chichilnisky.
\newblock The structure of large-scale synchronized firing in primate retina.
\newblock {\em Journal of Neuroscience}, 29(15):5022--5031, Apr 2009.

\bibitem{Shlens_JN_2006}
J~Shlens, G~D Field, J~L Gauthier, M~I Grivich, D~Petrusca, A~Sher, A~M Litke,
  and E~J Chichilnisky.
\newblock The structure of multi-neuron firing patterns in primate retina.
\newblock {\em J. Neurosci.}, 26(32):8254--66, 2006.

\bibitem{sohldickstein}
J~Sohl-Dickstein and B~Olshausen.
\newblock A spatial derivation of score matching.
\newblock {\em Redwood Center Technical Report}, 2009.

\bibitem{Tanaka:1998p1984}
T~Tanaka.
\newblock {Mean-field theory of Boltzmann machine learning}.
\newblock {\em Physical Review Letters E}, Jan 1998.

\bibitem{Welling:2002p3}
M~Welling and G~Hinton.
\newblock {A new learning algorithm for mean field Boltzmann machines}.
\newblock {\em Lecture Notes in Computer Science}, Jan 2002.

\bibitem{Welling:2003p7517}
M~Welling, G~Hinton, and S~Osindero.
\newblock Learning sparse topographic representations with products of
  student-t distributions.
\newblock {\em ADVANCES IN NEURAL INFORMATION PROCESSING SYSTEMS}, Jan 2003.

\bibitem{Yuille04}
A~Yuille.
\newblock The convergence of contrastive divergences.
\newblock {\em Department of Statistics, UCLA. Department of Statistics
  Papers.}, 2005.

\end{thebibliography}


\begin{thebibliography}{1}

\bibitem{macke}
J~Macke and S~Gerwinn.
\newblock Personal communication.
\newblock 2009.

\end{thebibliography}
}

\end{document}